\documentclass[manuscript]{acmart}

\AtBeginDocument{%
  \providecommand\BibTeX{{%
    \normalfont B\kern-0.5em{\scshape i\kern-0.25em b}\kern-0.8em\TeX}}}

\setcopyright{acmcopyright}
\copyrightyear{2022}
\acmYear{2022}

\acmConference[FAccT '22]{Woodstock '22: ACM Conference on Fairness, Accountability, and Transparency 2022}{June 21--24, 2022}{Seoul, South Korea}
\acmBooktitle{FAccT '22: ACM Conference on Fairness, Accountability, and Transparency,
  June 21--24, 2022, Seoul, South Korea}

\usepackage[ruled]{algorithm2e}
\usepackage{wrapfig}
\usepackage{tikz}
\usepackage{setspace}
\usepackage{enumitem}
\usepackage{titlesec}

\titlespacing*{\section}{0pt}{10pt}{0pt}
\titlespacing*{\subsection}{0pt}{8pt}{0pt}

\begin{document}

\title{Trust in AI: Interpretability is not necessary or sufficient, while black-box interaction is necessary and sufficient}
\renewcommand{\shorttitle}{Trust in AI: Interpretability is not necessary or sufficient, while black-box interaction is}

\author{Max W. Shen}
\email{maxwshen@gmail.com}
\orcid{0000-0002-4226-1861}
\affiliation{%
  \institution{Genentech}
  \streetaddress{1 DNA Way}
  \city{South San Francisco}
  \state{California}
  \country{USA}
  \postcode{94080-4990}
}




\begin{abstract}
  The problem of human trust in artificial intelligence is one of the most fundamental problems in applied machine learning. 
  Our processes for evaluating AI trustworthiness have substantial ramifications for ML's impact on science, health, and humanity, yet confusion surrounds foundational concepts. 
  What does it mean to trust an AI, and how do humans assess AI trustworthiness? 
  What are the mechanisms for building trustworthy AI? 
  And what is the role of interpretable ML in trust? 

  Here, we draw from statistical learning theory and sociological lenses on human-automation trust to motivate an AI-as-tool framework, which distinguishes human-AI trust from human-AI-human trust.
  Evaluating an AI's contractual trustworthiness involves predicting future model behavior using behavior certificates (BCs) that aggregate behavioral evidence from diverse sources including empirical out-of-distribution and out-of-task evaluation and theoretical proofs linking model architecture to behavior. 

  We clarify the role of interpretability in trust with a ladder of model access. Interpretability (level 3) is not necessary or even sufficient for trust, while the ability to run a black-box model at-will (level 2) is necessary and sufficient. 
  While interpretability can offer benefits for trust, it can also incur costs. We clarify ways interpretability can contribute to trust, while questioning the perceived centrality of interpretability to trust in popular discourse.

  How can we empower people with tools to evaluate trust? 
  Instead of trying to understand how a model works, we argue for understanding how a model behaves.
  Instead of opening up black boxes, we should create more behavior certificates that are more correct, relevant, and understandable. 
  We discuss how to build trusted and trustworthy AI responsibly with contract-aware model design, robustness testing, and favoring trust calibration over maximization.
\end{abstract}



\maketitle


\section{Introduction}
\label{section:intro}

Rapid advances in machine learning (ML) have spurred interest in artificial intelligence (AI) for increasingly diverse high-impact applications including medical decision making, self-driving cars, and generating text for propaganda and fake news. 
At the same time, deep models are increasingly understood to habitually engage in shortcut learning \cite{shortcut} and lack the ability to generalize in a systematic manner \cite{systematic}.
These concerns magnify when AI are in feedback loops: models minimizing average loss can perform worse on minority populations, the ensuing degraded service quality can deplete minorities from the service and future training data \cite{tatsu}. Flaws in deep-learning bioimage models in impactful publications can affect the training of future models and propagate to other papers \cite{bioimage}.
This state of the field demands us to consider how well we can trust AI -- can we trust AI to recommend useful treatments for never-before-seen patients, to act fairly with minimal bias, to preserve data privacy, to align with human values? 
Can we trust AI to not kill via self-driving cars, to not hinder scientific discovery, to make the world a better place?

Many have been alarmed by the black box nature of many state-of-the-art ML models, and hesitant to trust these models given our inability to understand what these models have learned and how they make decisions. 
Interpretability technology, including explainability, transparency, understandability, legibility, and intelligibility -- generally considered as the ability to understand the internal logic, inner workings, and rationale behind predictions -- are widely touted as a critical and necessary tool for trust. 
NIST, a U.S. national lab that aims to influence technology standards, writes that explainability is necessary to determine that an AI system is trustworthy \cite{nist}. 
An editorial in \textit{Nature Biomedical Engineering} writes "for trust... opening up algorithms to interpretation is a necessary first step" \cite{natbme}. In Google's AI best practices: "Interpretability is crucial to... trust AI systems." \cite{google}
"ML systems must be transparent to earn experts' trust" \cite{beenkimthesis}. 
"The demand for the ability to... trust ML systems [is increasing], for which interpretability is indispensable" \cite{carvalho}.
"It is very hard to trust any model ... without having transparency into how those models operate" \cite{schmelzer}.
The perception that interpretability is critical to trust is incredibly widespread \cite{lime, tonekaboni, rudin, debate, molnar, lipton2017doctor, deployment, yoonmedethics}. 
However, progress on interpretability has been difficult to measure, as lack of a clear consensus definitions have exposed interpretability's inherent subjectivity and field-specific meanings \cite{maxfield, krishnan, lipton, doshivelez2017rigorous, lipton2017doctor}. 
This has limited the real-world impact of interpretability methods \cite{bhatt}. 

Furthermore, a moment's reflection reveals that trusting black boxes is possible.
Air craft collision systems operate without human intervention \cite{doshivelez2017rigorous}. 
Doctors trust electrocardiogram machines without knowing how they work; in fact, a doctor that refuses black boxes "will be standing in the operating room with their stethoscope and little else" \cite{maxime, maxfield}.
Heart patients manually check their pulse after receiving a pacemaker, but gradually learn to trust their pacemaker as time goes on \cite{tschopp}. 
Best practices to avoid replication crisis from deep-learning-based bioimage analysis focus on robustness testing and performance validation on held-out data, not interpretability \cite{bioimage}.
Humans can also trust other humans in this fashion: 
bettors can trust chess experts to perform well based on past performance, without explanations of mental processes understandable only to other chess experts. 
The idea that black-box models do not preclude trust or usefulness has been argued on philosophical \cite{duran, krishnan} and pragmatic grounds \cite{holm}, including by doctors \cite{maxime}.

The tension between these observations and interpretability's asserted centrality to trust remains largely unresolved in literature, and
problematically,
the precise nature of the relationship between interpretability and trust remains  poorly studied.
If trust is a primary end-goal of interpretability, we must ask how precisely interpretability contributes to trust, and if there are better ways to advance trust.


\textbf{Overview.}
In section \ref{section:2}, we address what it means to trust AI, and how humans do so. 
We ground our definition of AI in statistical learning theory, avoiding an anthropomorphizing of AI inherent to other lenses on human-AI trust \cite{toreini, jacovi}.
By avoiding assigning moral agency to AI, we reduce the human-AI trust problem to concerns on the abilities and reliability of its input-output behavior on contracts, which we frame in terms of out-of-distribution and out-of-task performance. 
We later study human-AI-human trust, which can be decomposed into three problems: two of human-AI trust and one of human-human trust. 

Human-AI trust reduces to concerns on model behavior, and humans evaluate trust by gathering information on model behavior in what we call behavior certificates (BCs) via black-box interaction or interpretability.
We ontologize BCs into non-interactive and interactive BCs, and introduce a 5-point scoring system on their correctness, contract relevance, and understandability.
The academic field of ML research, and practical efforts in applied ML, can be viewed as systems for producing BCs. 

We then ask what interpretability contributes to trust. In section \ref{section:3}, we introduce a ladder of model access and insight relating interpretability and transparency to the ability to run a black-box model at-will (black-box interaction). 
In section \ref{section:4}, we answer: what is the minimum level of model access necessary for trust in AI? 
We conclude that black-box interaction is sufficient, thus interpretability is not necessary. We further argue that interpretability alone is not sufficient for trust, while black-box interaction is in fact necessary.

In section \ref{section:5}, we discuss how to build trustworthy AI, evaluate trust, and communicate trust. 
Rather than interpretable models, we suggest that understandable behavior certificates, allowing diverse audiences to understand and reason about model input-output behavior, are more useful. 
We also highlight the importance of clearly defining trust contracts, contract-aware model design, and pursuing trust calibration over maximization.


In section \ref{section:6}, we investigate the benefits and costs of pursuing level model interpretability or transparency. 
While not strictly necessary for trust, we highlight specific advantages: interpretability can enjoy substantial data efficiency and gracefully handle underspecified trust contracts, and produce more relevant and precise BCs.
However, interpretability can cost: second-order trust problems can arise when the trustworthiness of model explanations is unclear \cite{murdoch}, and weak transparent models can be less trustworthy than stronger black-box models. We also discuss interpretability's role in legal compliance, model debugging, and scientific discovery.


\section{The Problem of Human Trust in AI}
\label{section:2}

What do we mean when we ask whether humans can trust an AI? Here, we reduce human-AI trust to concerns on model behavior. To evaluate trust, humans gather information on model behavior by black-box interaction or interpretability, and use this evidence to reason about the model's potential behavior on trust contracts.

\textbf{Defining AI.}
In statistical learning theory, an ML model is a function $g$ that maps data from some input domain to some output domain, with parameters $\theta$ that are learned from training data \cite{Bousquet2004}. 
Typical AI systems can include ML components alongside other algorithmic processes $h$ (arbitrary data-transforming functions, i.e., data preprocessing or output postprocessing), all of which are executed by a Turing machine. 
Any functional composition $g \circ h$ is also a data-transforming function $f$, which we define as AI. 
AI's real-world impact stems from using $f$'s output data for decision making, in a human-mediated or automated manner, by machines or humans.
This definition of AI is sufficient to express automation systems such as medical decision trees, airplane autopilot systems \cite{hoffbashir}, foundation models \cite{foundation} and large language models \cite{gpt3}, superhuman reinforcement learning models like AlphaGo \cite{alphazero}, and captures the zeitgeist of modern and future fruits of machine learning research \cite{drexler}.

\textbf{Determinism.}
Without loss of generality, $f$ can be viewed as deterministic since stochastic models can be expressed by including random seeds as partial inputs. Model behavior is therefore space-time invariant and memoryless: we can run the computer program over and over, anywhere, anytime on the same input and $f$ will always produce the same output.
While the effective degree of determinism depends on our control over the model's input and the circumstances of its usage (e.g., if the model operates on stochastic input beyond our control, the model strays towards being nondeterministic to us), one can observe that AI behavior already differs substantially from human behavior. 
Human beings, relative to other human beings, effectively enjoy free will \cite{nichols} --
it is within the scope of human ability to realize the space-time invariance of AI behavior by running it repeatedly on identical input data, while the same is difficult to achieve with a human being. In practice, random seeds and input data are controllable in modern ML, rendering models effectively deterministic \cite{bouthillier}. This view supports the parallels between ML and software engineering, which can be viewed as two approaches with the same end-goal: creating computer programs \cite{mccarthy1998artificial}.


\textbf{Human-AI trust vs. human-AI-human trust.}
As a data-transforming function with deterministic behavior, AI is best understood as a tool. Like a double-edged sword, a deterministic data-transforming function is not a moral agent: it lacks inherent moral responsibility or agency \cite{sepresponsibility}.
Human-AI trust, then, becomes a vastly different problem than human-AI-human trust, where a human (i.e., a user) interacts with an AI that is wielded or deployed by another human (or a company) with potentially conflicting values and incentives \cite{jacovi}. 
In practice, human-AI-human trust problems are common, particularly challenging, and especially important, but it is critical to cleanly divide the two. Here, we focus primarily on human-AI trust, but briefly apply our findings to human-AI-human trust in section \ref{section:humanAIhuman}.

\textbf{Human-AI trust concerns deterministic behavior.}
Several recent treatments on human-AI trust begin from interpersonal (human-human) trust frameworks \cite{jacovi, toreini}. 
ABI+ is one influential framework in which human-human trust relies on ability, benevolence, integrity, and predictability \cite{abi, abiplus}. 
However, the anthropomorphization of AI in these analyses is incompatible with the determinism and lack of moral agency implied by statistical learning theory. 

AI should not be considered as a human being in trust transactions, but rather as an automation system. This view is supported by sociological research on human trust in automation systems, defined as "technology that actively selects data, transforms information, makes decisions, or controls processes" \cite{hoffbashir, leesee}. The inclusion of AI as a type of automation system is supported by recent E.U. GDPR policy \cite{gdpr, counterfactualexplanation}.
Empirically, humans also do not treat AI and automation systems like fellow humans in trust transactions \cite{madhavan}.
Human-human trust tends to transition from reliability to faith as familiarity increases \cite{leesee}, while human-automation trust follows a different arc over time, starting with an initial positivity bias, such that initial trust is based in faith and overly optimistic \cite{DZINDOLET2003697}, but as familiarity increases, the basis of trust transitions to performance and reliability \cite{madhavan, hoffbashir}. 

Another perspective on trust is Baier's notion of reliance for trust in inanimate objects (i.e., tools) \cite{baier}. We can understand reliance in terms of the ABI+ framework's concepts of ability and predictability, where 
ability describes the actions or behavior that a tool is capable of, and predictability is the consistency of behavior. The ABI+ concepts of benevolence (the trustee's desire to do good to the trustor) and integrity (the trustee's adherence to values or principles accepted by the trustor) are discarded as they fail to apply to non-moral agents. 
Thus, human-automation trust mainly depends on performance \cite{leemoray}.

Anthropomorphizing AI is cautioned against by ML experts \cite{Jordan2019Artificial}. Such attitudes can mislead trust judgments to public harm. It is critical that developers engaging with the public correct such misconceptions when possible.




\textbf{Contractual trust.}
Notions of trust are incomplete when what the trustee is trusted to do, or the type of behavior the trustee is trusted to have, is unspecified. 
To address this issue, we employ the framework of contractual trust \cite{jacovi}, or trust with commitment \cite{hawley, tallant}, where the trustee is trusted to fulfill a certain contract, which can concern any aspect of the model's behavior: correctness, performance, fairness, preserving data privacy, behaving ethically, resources used such as wall-clock time or memory, etc. 
The expressivity and generality of the contractual trust framework elevates trust to a centrally important role in applied ML by subsuming, like an umbrella, any model desiderata.

\textbf{Model behavior contracts concern out-of-distribution and out-of-task performance.}
To define contracts and quantify contract success, we can extend the statistical learning framework of training models on \textit{training data} and \textit{training tasks}. We propose to define tasks in full generality as computable functions of a model's behavior that return 'loss' values. 
Trust contracts 
can thus be expressed in terms of \textit{contract tasks}
and \textit{contract data distributions}. 

A key challenge arises when the training tasks and datasets are distinct from the trust contract tasks and data distributions.
In ML, the problem of out-of-distribution (OOD) generalization -- the model's performance on the training task on new data -- is widely studied, through lenses such as distributionally robust optimization \cite{rahimian2019distributionally}, invariant risk minimization \cite{ahmed2021systematic}, and more broadly through anomaly detection for abstaining or predicting with uncertainty. 
Model performance on tasks beyond the training task, which we call \textit{out-of-task} (OOT) performance 
is challenging to study because in general, performance on one task (e.g., minimizing squared regression loss) can have little bearing on performance on other tasks (e.g., fairness). 
Advances in meta-learning, few-shot, and zero-shot learning enable learning models that learn quickly on new tasks with limited data or no data at all \cite{maml, gpt3}. 

The rapid improvements in OOD and OOT performance in recent years, however, remain limited in the context of the panoply of tasks and data distributions desirable in trust contracts. 
With trust contracts, shifts in both dataset and task are common, though they are conventionally studied separately in ML. 
While models are often trained to minimize average loss over diverse datasets, each user's trust in the AI can concern its behavior on a unique distribution centered on data most relevant to each user \cite{tatsu}.
Users can also demand unique preferences on model desiderata -- not only through personal trade-offs among high-level notions like fairness, data privacy, and performance; but individual concepts like fairness can be defined and quantified in dozens of distinct ways, some of which are contradictory \cite{narayanan}.
Thus, while ML research promises to advance OOD and OOT performance, many more challenges must be faced to advance trustworthiness.

\textbf{Contract expressivity.}
Over the preceding sections, we have argued that all human-AI trust problems can be expressed in terms of desired quantitative performance on out-of-distribution generalization or out-of-task performance. 
However, some desiderata such as adherence to ethical or moral principles can be challenging to express as computable functions of a model's behavior, which have motivated some to favor interpretability \cite{lipton, doshivelez2017rigorous}. 
Instead, we advocate for the other fork in the road: instead of abandoning the effort to clearly define our trust desiderata in quantifiable, computable terms, we believe that overcoming these challenges is crucial for human-AI trust. 

\subsection{Formalizing the problem of human-AI trust}




An AI is a deterministic data-transforming function $f$ (which we can think of as a model or a computer program) that maps data from some input space to some output space. 
A task $\mathcal{T}$ is a computable function of $f$ and some input data, which measures some property of $f$'s behavior by running $f$ on the input data. 
In ML, models are trained
on some training task $\mathcal{T}_{\texttt{tr}}$ on a training dataset $\mathcal{X}_{\texttt{tr}} \sim \mathcal{P}_{\texttt{tr}}(\mathcal{X})$.
A \textit{trust contract} $C = (\mathcal{P}_{\texttt{con}}(\mathcal{X}), \{ \mathcal{T}_{\texttt{con}}^i \},  \mathcal{S}_{\texttt{con}} )$ includes a contract data distribution $\mathcal{P}_{\texttt{con}}(\mathcal{X})$, a set of contract tasks $\{ \mathcal{T}_{\texttt{con}}^i \}$, and a success evaluation function\footnote{It is convenient to assume that $\mathcal{S}_{\texttt{con}}$ can be defined to operate either on the distribution $\mathcal{P}_{\texttt{con}}(\mathcal{X})$ or on samples $\mathcal{X}_{\texttt{con}} \sim \mathcal{P}_{\texttt{con}}(\mathcal{X})$} $\mathcal{S}_{\texttt{con}}(f, \{ \mathcal{T}_{\texttt{con}}^i \}, \mathcal{X}_{\texttt{con}} ) \rightarrow \{ 0, 1 \}$ that identifies $f$'s performance on the contract tasks and contract data distribution as either a failure or success.

The \textit{contractual trustworthiness problem} is: given a model $f$ and a trust contract $C$, output a contractual trustworthiness score, representing our subjective likelihood\footnote{This reflects other human trust processes that encompass both objective and subjective elements \cite{sep-trust}.} that the model will succeed on the contract, using evidence about model behavior available to us.
These evidence are \textit{behavior certificates}, denoted by $\mathcal{B}$, which can regard any aspect of model behavior, even those not relevant to the model's likely behavior on the trust contract. 
The \textit{contractual trustworthiness score} can thus be written as\footnote{Excitingly, estimating or inferring contractual trustworthiness scores for a model can itself be solved with ML. At present, ML models may struggle with the diverse modalities and uncertainties in behavior certificate evidence. Key challenges also include communicating human-discovered BCs to models and ML-guided discovery or generation of BCs.}:


\begin{equation}
  P
  \Big(
    \mathcal{S}_{\texttt{con}}(
      f, 
      \{ \mathcal{T}_{\texttt{con}}^i \}, 
      \mathcal{P}_{\texttt{con}}(\mathcal{X})
    )
    = 1
    \Big| 
    \mathcal{B}
  \Big)
\end{equation}


\begin{figure}
  \usetikzlibrary{arrows}
  \tikzstyle{required} = [rectangle, rounded corners, minimum width=2.5cm, minimum height=0.5cm, text width=2.5cm, text centered, draw=black]
  \tikzstyle{optional} = [rectangle, rounded corners, minimum width=2.5cm, minimum height=0.5cm, text width=2.5cm, text centered, draw=black, fill={rgb:black,1;white,7}]
  \tikzstyle{arrow} = [thick,->,>=stealth]
  \tikzstyle{optional arrow} = [thick,->,>=stealth]

  \begin{tikzpicture}[node distance=1.2cm, font=\small]

    \node (model) [required] {Trained model with access level 1, 2, or 3};
    \node (ood) [optional, below of=model] {\textit{Additional data}};
    \node (oot) [optional, below of=ood] {\textit{Additional tasks}};
    \node (contract) [required, right of=model, xshift=3cm] {Trust contract, specific desires on model behavior};
    \node (bcs) [required, below of=contract, yshift=-0.5cm] {Behavior certificates, information on any aspect of model behavior};
    \node (score) [required, right of=contract, xshift=4.5cm, yshift=-0.5cm] {Contractual trustworthiness score};

    \draw [arrow] (model) -- (bcs);
    \draw [dashed, ->] (ood) -- node[anchor=north] {\textit{Optional}} (bcs);
    \draw [dashed, ->] (oot) -- (bcs);
    \draw [arrow] (contract) -- node[anchor=north, yshift=0.7cm] {Estimation / Inference} (score);
    \draw [arrow] (bcs) -- (score);
  \end{tikzpicture}

  \caption{The human-AI contractual trustworthiness problem}
  \label{fig:overview}
\end{figure}
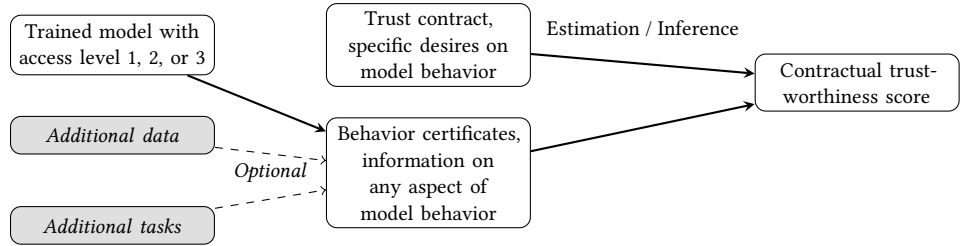

\textbf{Estimation and inference.} 
In Fig. \ref{fig:overview}, a human has access to a trained model, and defines a trust contract expressing specific desires on model behavior that the human would like to trust the AI on. To evaluate the AI's contractual trustworthiness, the human collects information on any aspect of model behavior they would like in the form of behavior certificates, which are used as evidence to reason about the model's potential behavior on the trust contract.

When the contract is accessible -- we have abundant independent samples from the contract data distribution, and contract tasks are clearly defined -- we can directly compute evidence, or behavior certificates, regarding the model's behavior on the contract. This embodies an \textit{estimation} problem in the statistical sense of calculating a property value of a population from samples $\mathcal{X}_{\texttt{con}} \sim \mathcal{P}_\texttt{con}(\mathcal{X})$. 
When the sample size $N$ is large, it is especially powerful to train the model directly on the contract task and data; models can be trained on contract tasks even when they are not differentiable functions of model parameters, using techniques such as score function estimators \cite{reinforce}, straight-through gradient estimators \cite{ste}, and evolutionary algorithms \cite{evolutionary}. After training, performance on held-out data then serves as a powerful behavior certificate estimating contract adherence with small error bars. 
When $N$ is small, models can be trained on surrogate data\footnote{Sometimes, also surrogate tasks}, and evaluated on contract data to statistically estimate contract behavior with larger error bars. 

In other cases, the contract is not accessible: the contract data distribution or contract tasks may be underspecified, or we may not have access to contract data samples before deployment. In these situations, the trust problem is solved by \textit{inference} in the statistical sense of reasoning about a latent variable from observed evidence. Without the ability to evaluate the model on the contract, we use behavior certificates and surrogate tasks and data to infer the model's potential behavior on the contract.



\textbf{Acting on Contractual Trustworthiness Scores.} Individual contracts, when defined with proper precision, can be narrow. Complex real-world applications likely demand many contracts, each corresponding to particular model desiderata such as robustness or fairness. A model could be very trustworthy on one contract but not others.
The trustworthiness scores for many contracts should be combined with analyses of risk, potential consequences, and other factors, to decide the level of human involvement in the AI's decision making. The 10-level Sheridan-Verplanck scale of automation \cite{sheridanverplanck} ranges from human oversight, varying degrees of human involvement in the loop, to full independence.

\subsection{Behavior certificates}
\label{section:bc}

Our reduction of human-AI trust to concerns on model behavior elevates the importance of our processes for collecting evidence on model behavior, and evaluating and reasoning about behavior evidence. 
We clarify these processes with the concept of \textit{behavior certificates} (BCs), which are (evidence, conclusion) tuples; pieces of evidence supporting some conclusion about a model's behavior. 
Our notion of behavior certificates generalizes\footnote{While Jacovi's framework places trust as the central output, our central output is understanding of model behavior, which then leads to trust.} Jacovi's causes of trust in AI \cite{jacovi}.
As the behavior described by a BC regards a computable property of a model, BCs correspond explicitly or implicitly to some task $\mathcal{T}_{\texttt{bc}}$ on some data $\mathcal{X}_{\texttt{bc}}$. 
To understand their central role in trust, we provide an ontology of BCs, discussing two main types, their provenance and key properties, and examples.

\textbf{Interactivity and provenance.} 
\textit{Interactive BCs} are formed by interacting with a given pre-trained model, typically by running it on new data or tasks. 
They are often constructed and discarded on a per-model and per-problem basis. When trust contract tasks or data are not available, interactive BCs of model behavior on surrogate data or tasks can help us reason or infer about the model's possible behavior on a trust contract. Interestingly, the construction of interactive BCs may be scaled with automation: one can imagine using generative models to produce and label data, running models on established collections of useful tasks, or tasks learned through methods like inverse reinforcement learning, and using automated adversarial methods to identify pertinent adversarial examples.

\textit{Non-interactive BCs} can be formed solely from model design and training details, such as loss function, regularizers, model architecture, and data available during training (training, validation, and test sets). These BCs can describe the impact of certain architecture choices or regularization strategies on the behavior of the final trained model.
The production of novel non-interactive BCs often requires academic research and human creativity, and are typically long-lasting and broadly relevant to many models and problems. They are often collected by developers in conceptual toolboxes to help design ML approaches for new problems. 

\noindent
\textbf{Examples of Non-interactive Behavior Certificates}
\vspace{-3pt}
\begin{itemize}[leftmargin=10pt]
  \item \textit{(In-distribution validation)} The held-out validation loss is 1.2. $\Rightarrow$ The model's expected loss on $\mathcal{P}_{\texttt{tr}}(\mathcal{X})$ is 1.2.
  \item \textit{(Geometric deep learning \cite{bronstein2021geometric})} The model comprises self-attention layers without position embeddings followed by mean aggregation. $\Rightarrow$ The model is permutation invariant: for any data $x$, $f(x) = f(\texttt{permute}(x))$.
  \item \textit{(Data augmentation)} The image classification model was trained with rotation and color data augmentation. $\Rightarrow$ For any data $x'$, $f(x') \approx f(\texttt{rotate}(x')) \approx f(\texttt{color}(x'))$ under the same perturbation distributions used during training.
  \item \textit{(Models designed with prior knowledge)} The model is a universal differential equation \cite{rackauckas} incorporating knowledge of intraspecies competition. $\Rightarrow$ On any data, the learned dynamics are consistent with intraspecies competition.
  \item \textit{(Regularization)} A model is trained to a distributionally robust optimization regularizer value of 1.6 with distance $r$. $\Rightarrow$ Model loss is guaranteed to be 1.6 or lower on any data distribution within $\chi^2$-divergence of $r$ to $\mathcal{P}_\texttt{tr}(\mathcal{X})$. \cite{tatsu, rahimian2019distributionally}.
  \item \textit{(Interpretability)} Feature importance analysis finds that the pre-trained model uses spoons to classify images labeled 'chocolate sauce'. $\Rightarrow$ The model is likely to incorrectly classify images with spoons and without sauce as chocolate sauce, and misclassify images of chocolate sauce without spoons as not chocolate sauce \cite{fong}.
\end{itemize}

\noindent
\textbf{Examples of Interactive Behavior Certificates}
\vspace{-3pt}
\begin{itemize}[leftmargin=10pt]
  \item \textit{(OOD evaluation / Robustness testing)} A pre-trained model has loss of 2.1 on surrogate data $\mathcal{X}_\texttt{bc}$. $\Rightarrow$ The model has expected loss of 2.1 over $\mathcal{P}_\texttt{bc}(\mathcal{X})$ with $\mathcal{L}_{\texttt{tr}}$.
  \item \textit{(OOT evaluation)} A model trained with cross-entropy loss is evaluated on a fairness task $\mathcal{T}_\texttt{fair}$ and performs poorly. $\Rightarrow$ The model is not fair (as defined by $\mathcal{T}_\texttt{fair}$) on $\mathcal{P}_{\texttt{tr}}(\mathcal{X})$ and likely on other data.
  \item \textit{(Adversarial examples)} We find, using human intuition or an automated method, a rotated version of a training image that the model classifies incorrectly, despite being correct on the unrotated image. $\Rightarrow$ The model is not rotation invariant, and will likely misclassify rotated versions of other images \cite{engstrom2019exploring}.
\end{itemize}

\textbf{Grading behavior certificate quality.} BCs can vary in their relevance to the contracts, their correctness, and understandability, and each aspect can be rated on 5 point scale. For instance, some BCs backed by mathematical proofs can have near-perfect correctness (5), but have low relevance to contracts of interest (1-2), and be less understandable to laypeople (1-2). Meanwhile, some BCs formed by held-out validation can have high correctness (4), strong relevance (4), and are easier to understand (4). In this light, an important area of research in trust is improving our ability to generate BCs with greater correctness, relevance to contracts, and understandability. 

\textbf{Correctness.} Correctness regards how well the conclusion follows from the evidence, which can be surprisingly challenging without expert knowledge in both ML and application domains. For instance, a common pitfall of early deep medical models trained on hospital A's data distribution $\mathcal{P}_\texttt{A}(\mathcal{X})$ was erroneously concluding that the model will perform well on hospital B; the conclusion should have been more narrow, as the evidence only supports good performance on $\mathcal{P}_\texttt{A}(\mathcal{X})$ \cite{zech, futoma}. Correctness is improved by forming conservative and precise conclusions that are sensitive to data assumptions and possible batch or environmental effects in the data. 

\textbf{Relevance to contracts.} Relevance can be improved by creating BCs on contract tasks and data. 
Distribution testing can assess contract relevance of held-out validation statistics: if $x_{\texttt{con}} \sim \mathcal{P}_{\texttt{bc}}(\mathcal{X})$, then our interactive BCs are strongly relevant to that contract.
\textit{Contract-aware model design} with inductive biases and domain knowledge can help construct non-interactive BCs with improved contract relevance. 

\textbf{Understandability.} Understandability concerns how well we can understand the evidence to be true, understand how the conclusion derives from the evidence, and understand the statement of the conclusion and its implications on model behavior. 
Understandability requires case-by-case effort in scientific and technical communication. 
We propose that BCs should generally be communicated in multiple ways: in technically precise language, for fellow developers, but also in plain language, for laypeople. This empowers a diverse audience to reason about model behavior and trustworthiness, while handling the trade-off that understandability to laypeople often introduces ambiguity for experts. 
Summarizing BCs in model cards can aid understandability \cite{modelcards}.

\subsection{The Problem of Human-AI-Human Trust}
\label{section:humanAIhuman}

The problem of human-AI-human trust is considerably more complex than human-AI trust, as it can be decomposed into three trust problems: two human-AI trust problems, and a human-human trust problem. In our AI-as-tool lens, this problem manifests when a human interacts with an AI that is wielded as a tool by another human. The importance of the problem of human-human trust to applied ML cannot be overstated, as AI applications with ever-greater ability and impact are deployed by humans and companies operating by diverse incentives and values. While a proper treatment of this problem is beyond the scope of this paper, we offer a brief analysis here, using our human-AI trust framework. Regulation and transparency into engineering practices will be crucial for building human-AI-human trust.

\textbf{Two human-AI trust problems.} Issues can arise from misalignment in perceived trustworthiness levels. Consider an example where a developer releases a product that grants an AI a high Sheridan-Verplanck automation level based on the developer's trust in the AI to be accurate and fair in real-world applications. The user, having seen a different body of evidence and BCs, agrees with the developer's trustworthiness score in the AI's ability to be accurate, but assigns lower trustworthiness in the AI's ability to be fair. Overall, the user prefers a lower automation level than provided, but this is outside of their control. Left with only the choice of either using the product or not, they choose to not use it.

In general, disagreements at any point in the information flow in Figure \ref{fig:overview} can deteriorate human-AI-human trust. If information or evidence is asymmetrically accessible, it should be unsurprising that trust conclusions are likely to disagree. If evidence from BCs are publicly shared, humans may still disagree on the conclusions derived from the evidence, or on the relevant tasks or data distributions. And even if people agree on contractual trustworthiness scores, they may disagree on the potential consequences and risks in deploying the AI.

\textbf{Human-human trust.} Human-human trust can be positive: strong interpersonal trust can enable users to bypass the difficulties of assessing human-AI trust while enjoying the benefits of AI use. 
It can also
introduce incentive and value alignment challenges. Even if both parties share strong human-AI trust in sharing similar trust in the behavior of the AI, lack of human-human trust can manifest as a lack of trust in each other's uses of the AI, analogous to trust issues with nations (moral agents) and nuclear weapons (tools). Nations can trust that the other nation's nuclear weapons work as advertised, but may not trust how the other nation will use their nuclear weapons.

\section{The Ladder of Model Insight}
\label{section:3}


We now turn to the question: what role does interpretability et al.\footnote{Interpretability et al.: transparency, explainability, understandability, intelligibility, legibility, etc} play in human-AI trust? 
Before embarking on such analysis, we first introduce a ladder of model access and insight to relate interacting with black-box models with interpretability et al. which "open up the black box". 

The \textit{ladder of model insight}\footnote{We take inspiration from Judea Pearl's Ladder of Causality. \cite{pearl}} contains three levels of questions with increasing detail (Figure \ref{fig:access}). 
\textit{Level $x$ insight} is information obtained from answers to level $x$ questions. 
\textit{Level $x$ access} is the general amount of access to the model needed to accurately\footnote{Accuracy is important: explainability methods use level 2 access to approximate answers to level 3 questions, but if they are not accurate or faithful to the model's inner workings, then they have not produced level 3 insight} answer level $x$ questions\footnote{Reported model access levels describe access that is generally necessary to accurately answer questions of each level, based on the current state of the field at the time of publication. The access needed for level x questions may change over time as research progresses. Our definitions accommodate this possibility: the level of any question never changes, but the meaning of level $x$ access would change.}. 
We call this framework a ladder because, in typical settings, the ability to answer questions of a certain level implies the ability to answer questions of lower levels.
Model access typically accumulates over three main axes: awareness that the model exists, the ability to run a model which may be a black box, and the ability to understand the internals of a model. 

\begin{itemize}[leftmargin=10pt]
  \item \textit{Level 0 model access}: The human is not aware of the model, which may be making decisions for or about the human.
  \item \textit{Level 1 model access}: The human is aware of the model but is limited in their ability to run the model. Level 1 questions include giving examples of model input and output, and reasoning about the space of input and outputs. This can reveal what information the model may depend on and the kinds of decisions the model's output might be used for.
  \item \textit{Level 2 model access}: This level includes the typical access afforded to engineers or developers building black-box models: they can run the model at will, but cannot see inside the black box. Level 2 questions regard the model's performance on any task and dataset. 
  \item \textit{Level 3 model access}: Humans can run the model and see inside the model, which is considered transparent or interpretable under conventional definitions. With level 3 model access, we can answer level 3 questions regarding what the model has learned, how the model makes its decisions, and any other questions considered answerable with models considered to be interpretable et al. 
\end{itemize}


\begin{figure}
  \begin{center}
    \small
    \begin{tabular}{ c p{5cm} l l p{3cm} }
    \hline
    \multicolumn{2}{c}{} & \multicolumn{3}{c}{\textbf{Model access}} \\
    \cline{3-5}
    \textbf{Level} & \textbf{Questions} & \textbf{Aware of model} & \textbf{Can run model} & \textbf{Model clarity} \\
    \hline
    0 & None & \textcolor{lightgray}{No} & \textcolor{lightgray}{No} & \textcolor{lightgray}{Black box} \\
    1 & What is an example input and output? What is the space of the input and output? & Yes & \textcolor{lightgray}{Limited} & \textcolor{lightgray}{Black box} \\
    2 & How does the model perform on any task and dataset? & Yes & At will & \textcolor{lightgray}{Black box} \\
    3 & What has the model learned? How does the model make its decisions? & Yes & At will & Interpretable et al. \\
    \hline
    \end{tabular}
  \end{center}

  \caption{Summary of the Ladder of Model Insight}
  \label{fig:access}
\end{figure}

Model access is fluid: a researcher enjoys level 2 or 3 access when developing their own model, but only has level 1 access when interacting with a closed-source research model through a rate-limited web app. Laypeople can enjoy level 2 access to machine learning services like Google translate, but have level 0 access by lacking awareness of their participation in other machine learning-driven ad services by the same company. 

\section{What is the Minimum Level of Model Access Necessary or Sufficient for Trust?}
\label{section:4}




\textbf{Level 2 model access is sufficient to evaluate trust in AI.}
This result concisely follows from definitions in preceding sections. To estimate or infer contractual trustworthiness, it suffices to have the power to run the model on arbitrary tasks or data. This power is achieved with level 2 model access; Q.E.D. Note that the ability to evaluate trust does not guarantee trust in any particular model. If the model is inherently trustworthy, however, then it suffices for humans to come to trust an AI. 

A corollary of this result is that \textbf{level 3 model access is not necessary for evaluating trust in AI.} As level 3 access is level 2 access plus the ability to generate level 3 model insight (interpretability et al.), we can conclude that 
\textbf{interpretability et al. is not necessary for trust when we can run a model at will, even if the model is a black box.} This result explains a variety of observations on trust without interpretability et al. in the introduction.

While level 2 access is in principle sufficient, in practice, enormous effort can be required before trust is achievable. While many challenges can be addressed without interpretability (section \ref{section:5}), behavior certificates from level 3 model insight can sometimes be useful (section \ref{section:6}).

\vspace{2mm}
\noindent
\textbf{On many real problems, level 3 model insight is not sufficient for trust, while level 2 insight is necessary.}

How does interpretability et al. relate to trust if we cannot run the model at will? 
While this scenario is rare in practice\footnote{More precisely, it is rare that \textit{no one} can run a model at will while \textit{someone} can interpret it. The inability of any single person to run a model at will while interpreting it is an artificial constraint on model access; in principle, a person who can interpret the model can almost always be granted the ability to run the model at will}, analyzing such a question helps us further understand the relationship between interpretability et al. and trust. 
Without the ability to run the model at will, we lose level 2 insight and retain only level 3 insight.

In full generality, we can assess contractual trustworthiness as long as we can generate any type of behavior certificates, which means that level 3 insight alone can be used to evaluate trust. 
When models and trust contracts are simple, internal knowledge can suffice to accurately assess contract adherence and induce trust in a trustworthy model. 

However, the power of level 3 insight can scale poorly as models or contracts gain in complexity, or as applications gain in importance, while level 2 insight can become increasingly necessary. 
Software engineers can trust simple programs solely from reading code, but as programs increase in complexity, this approach is no longer sufficient as best practice begins to require unit tests and empirical validation of program behavior -- as Donald Knuth warns, "beware of bugs in the above code; I have only proved it correct, not tried it". 
Indeed, programs can be written that are impossible to trust from inspection alone \cite{trusting_trust}, and extensions of Gödel's incompleteness and Church-Turing's undecideability theorems tell us that programs exist where questions about their behavior are unanswerable without running it \cite{rice}.

More generally, trusting a model based on level 3 insight alone is analogous to trusting a system, or a scientific theory or model, or a medical drug, based on theoretical arguments over empirical demonstrations -- "In theory, theory and practice are the same. In practice, they are not." 
Theoretical arguments are not seen as sufficient.
Our practices of experimental validation of competing scientific theories and randomized controlled trials speak to humankind's elevation of empirical demonstration as a necessary gold standard of evidence.

\section{Insights on Building Trusted and Trustworthy AI}
\label{section:5}

We provide a brief overview of three areas: building trustworthy AI from scratch, trust transactions with pre-trained models, and garnering human trust in pre-trained AI whose trustworthiness is fixed. Central are three main ideas:

\vspace{-2mm}
\begin{itemize}[leftmargin=10pt]
  \item \textbf{Define trust contracts clearly.} Trust is only meaningful if we know what we are trusting the AI to do \cite{jacovi}. This can require both ML and domain expertise.
  \item \textbf{Prefer estimation over inference when evaluating contractual trust.} Interactive behavior certificates formed by evaluating models directly on trust contracts are the most powerful evidence for contractual trustworthiness. 
  \item \textbf{Acknowledge shortcut learning.} Have pessimistic prior beliefs on an AI's abilities \cite{shortcut}.
\end{itemize}

\subsection{Building trustworthy AI.} 
The trustworthiness of AI on any contract depends on how well the AI's abilities  fulfill the contract's demands. Broadly speaking, the recent progress in ML performance reflected by superhuman or state-of-the-art achievements in many domains is perhaps the most powerful force in improving the potential trustworthiness of AI on complex and difficult contracts demanded by high-impact applications.

At the same time, trust contracts are often an after thought in modern practice: training tasks and data distributions can differ markedly from contract or deployment tasks and distributions, with degraded performance due to shortcut learning.
When building AI from scratch, its contractual trustworthiness can be improved by \textbf{contract-aware model design} to obtain models that perform better on trust contracts. 
The simplest approach is training or finetuning models directly on trust contract tasks and data, though to create models that are trustworthy to each user, we need to acknowledge the end-user diversity of trust contracts.
Ideas like distributionally robust optimization to minimize worst-case loss over any subset of the data population rather than minimizing average loss can be important for building models that succeed on this diversity of trust contracts, and thus are more trustworthy to more people \cite{rahimian2019distributionally, tatsu}.

Incorporating prior knowledge as inductive biases in deep models is a powerful approach for contract-aware model design where a combination of ML and domain expertise can make large contributions \cite{kaggle}. 
For instance, including a physics engine in a reinforcement learning model \cite{diffphysics} guarantees that the model will make decisions in a manner consistent with described physics, while it can be extremely difficult to trust a fully-connected neural network to be 100\% consistent with known physics. 
Adding truth to a model can only help.
While the successes of end-to-end learning are admirable -- models can outperform humans at feature engineering -- it can also require enormous amounts of data. 
When trust contracts are critical and data is a bottleneck, designing models with inductive biases and including various kinds of prior knowledge can improve contractual trustworthiness: end-to-end models are not the end-all be-all \cite{e2e}.

\subsection{Trust transactions with pre-trained models.}
In applied ML projects, it is increasingly common to use or adapt a pre-trained model over training from scratch \cite{bioimage}. 
In such situations, we have level 2 model access, but may be uncertain about the model's abilities, which we interrogate by forming non-interactive BCs from studying the model's design and training process, and forming interactive BCs from OOD and OOT evaluation. 
Our recommendations extend the best practice of robustness testing \cite{djolonga2020robustness, robustness, bioimage} by placing out-of-task evaluation as a first-class citizen. 
Lessons from robustness testing and benchmarking transfer: the effort in properly evaluating model behavior should not be underestimated -- significant resources may need to be spent for evaluations with proper relevance and statistical significance \cite{bowmandahl}.

Shortcut learning, or the tendency of models to rely on shortcuts in decision-making that perform well during training but can fail to generalize, reflects the practical wisdom that models optimize "by the letter, not the spirit" of loss functions \cite{shortcut}. This wisdom encourages us to \textbf{have pessimistic prior beliefs on model abilities}, and to favor evidence-based data-driven discovery of model capabilities. 

\subsection{Human trust in trustworthy AI.}
Once the abilities of an AI are fixed, its trustworthiness is also fixed, but different people's trust in the AI can vary by their access to evidence and their understanding of the model. 

\textbf{Aim for trust calibration, not maximization.}
Issues arise when the level of trust (the trustor's attitude) mismatches the true level of trustworthiness (the property of the trustee). Such \textit{unwarranted trust} \cite{jacovi} manifests as under-trusting or over-trusting. 
Assigning Sheridan-Verplanck automation levels that are too high can be blamed for the fatal crash of Turkish Airlines Flight 1951, where the pilots continued to rely on autopilot after the failure of an altimeter, while assigning automation levels too low can be blamed for the fatal wreckage of the Costa Concordia cruise ship when the captain manually diverged from the computer navigation route only to hit a shallow reef \cite{hoffbashir}. 

To better calibrate each user's trust to the model, developers should ensure that users clearly understand the limits of model abilities: 
while it is natural to highlight strengths, we have a responsibility to also communicate weaknesses. 
When only presented with strengths, uninformed users may incorrectly extrapolate model abilities to unproven tasks and data, which can be combatted by carefully communicating pessimistic prior beliefs due to shortcut learning, uncertainty modeling, and distribution tests.
Developers also have a responsibility to correct misconceptions that distort human-AI trust transactions, including actively discouraging unwarranted anthropomorphization of AI.

\textbf{Understandable BCs, not interpretable models.}
Behavior certificates are central to trust, as contractual trustworthiness cannot be evaluated without evidence of behavior. In contrast, model interpretability is more peripheral, and its link to trust is entirely mediated by behavior certificates (Fig. \ref{fig:overview}). Thus, if we must choose one, we should prefer to make behavior certificates interpretable or understandable, not models.

Constructing BCs that are relevant to trust contracts, correct, and communicating them clearly can require substantial ML and domain expertise, whether the model is interpretable or not. 
It can be unproductive to solely rely on end-users to construct BCs -- end-users may not know what to do with an interpretable model. Instead, developers should accept this responsibility, and present pre-crafted and selected BCs to end-users.
This strategy is analogous to the division of labor between lawyers and juries, where lawyers construct arguments by presenting evidence in line alongside a technical understanding of the law, and juries use the evidence to make their own conclusions. 

A benefit of pursuing understandable BCs over interpretable models is that research progress may be easier to measure. Progress in interpretability has notoriously been difficult to track \cite{doshivelez2017rigorous}, and its definition varies by field, which is unsurprising in the light of its complex relationship with ML and domain expertise and individually formed BCs. 
In contrast, when BCs are presented clearly in terms of evidence and conclusions, their correctness, relevance, and understandability are easier to measure. 
A BC-centric perspective may even provide new tools for measuring interpretability based on the quality of BCs generated by users who can only inspect the model without running it.

\textbf{Elevate user levels of model access.}
Alongside understandable BCs, empowering users to form their own understanding of model behavior can also improve human trust in AI while also addressing some human-human trust challenges. While this is especially important for audiences with sufficient ML and domain expertise to leverage level 2 and 3 model access to form BCs, significant improvements in trust for end-users can be gained by elevating them from level 0 to 1. 
Beyond independent evaluation of model behavior, providing model access can help end-users clarify and define the trust contracts most important to them. 

\section{Benefits and Costs of Pursuing Level 3 Insight}
\label{section:6}

Despite not being sufficient or necessary for trust, interpretability et al. is not useless. On the contrary,
interpretability and black-box interaction enable constructing different kinds of BCs which can strengthen and complement each other in evaluating and building trust. 
At the same time, pursuing level 3 insight can also incur costs.

\subsection{Benefits of interpretability et al.}


While empirical evaluation of model behavior through black-box interaction is a powerful tool for trust, it requires clearly defining trust contracts and gathering sufficient data before any BCs can be made. These prerequisites can be demanding. 
In contrast, interpretability et al. enables generating BCs with improved data efficiency, requiring less or even no data, and can gracefully handle underspecified trust contracts. These advantages particularly shine in forming negative BCs of model limitations.
Compared to non-interactive BCs constructed with only level 2 model access, level 3 model access can construct BCs that are more relevant to many trust contracts, as they can describe the final trained model behavior more narrowly and precise. 

\textbf{Data efficiency.} 
This point can be motivated by considering a puzzle: what is the point of a model being human simulatable (an important subcategory of interpretable models \cite{lipton}), if computers can simulate models faster and more precisely? 
By reasoning about the model in a broader context including vague goals and hypothetical data, humans can hypothesize highly relevant inputs and outputs that form powerful BCs about model behavior, without access to real data.
While adversarial ML methods attempt to automate this process, they can struggle more than humans to ensure adversarial examples are realistic enough to be relevant in real use cases. In general, this is a marked advantage of interpretability over black-box interaction, where data collection demands can be onerous \cite{bowmandahl}.


\textbf{Handling underspecified trust contracts.} 
With an understanding of how the model works and what it has learned to do, humans can also understand model behavior to evaluate trust in an emotionally meaningful manner, albeit poorly defined, in the absence of well-specified trust contracts. 
For instance, a human may conclude a model is unfair and only later pinpoint the precise definition of fairness that is violated by the model. 
This can be useful, particularly when clearly defining trust contracts is a challenging process. 

However, using model behavior to define trust contracts inverts the responsible approach of first defining trust contracts, then building a model to adhere to it (Section \ref{section:5}). 
Attempting to salvage a pre-trained model by searching for its utility errs by prioritizing the hammer over the nail.
In addition, such models may be less trustworthy than models developed with the foresight and benefit of contract-aware model design.

\textbf{Negative behavior certificates.} 
Interpretability et al. can be especially useful for generating negative BCs that describe failure points and limitations of model behavior -- if a model reasons about a phenomenon in an incorrect way, then we know it is untrustworthy on any data input for that phenomenon. 
For instance, the observation that a linear model is used to model non-linear input-output relationships is a powerful negative BC. 
The evidence that a model uses spoons to classify images of chocolate sauce, because spoons are more common than jars in ImageNet examples of chocolate sauce, is a powerful negative BC that the model is poor at classifying chocolate sauce images without spoons.
Negative BCs can play an important role in curtailing excessive AI reliance and calibrating trust downwards. 

The power of negative BCs contrasts with the difficulty of certifying that a model will perform well on a trust contract, even if we have dozens of insights that the model learned the "right things" or reasons about the data in a manner we expect. For establishing positive trust, such armchair arguments are widely considered insufficient compared to empirical validation by black-box interaction (Section \ref{section:4}). 

\subsection{Issues with interpretability et al.}

There are two main approaches for interpretability: using inherently transparent models, and obtaining post-hoc explanations of trained models that may be black boxes \cite{molnar, rudin}. Each approach carries risks.

\textbf{Explanations incur additional trust problems.}
As Rudin writes, "Explanations must be wrong. They cannot have perfect fidelity with respect to the original model... otherwise the explanation would equal the original model" \cite{rudin}, mirroring the adage that all models are wrong, but some are useful.
Explanations can be unfaithful and inaccurate to the actual inner workings of the model, unstable, uninformative, contradictory, nonsensical, or vulnerable to adversarial attack \cite{rudin, molnar}. 
When using model explanations, we then not only ask whether we can trust the model, but also whether we can trust the explanations: this presents a second trust problem \cite{murdoch}.

\textbf{Weak transparent models are less trustworthy than strong black-box models.}
Inherently transparent models are not necessarily weaker than black-box models, and can be useful for important tasks such as criminal recidivism prediction \cite{rudin}.
However, on many other high impact applications of ML, \cite{bioimage, gpt3}, state-of-the-art performance has not been achieved by transparent models. 
As contractual trustworthiness is inherently linked to a model's ability to succeed on a contract, if it is clear that a black-box model succeeds on a trust contract better than a transparent model, then the black-box model is more trustworthy.

\textbf{Explanations \& transparency can increase trust unwarrantedly, reducing calibration.}
Empirical studies has found that humans can err by placing too much trust in transparent models, assigning them higher Sheridan-Planck automation levels than black-box models even when the transparent model's performance was lower \cite{poursabzisangdeh2021manipulating}. 
Humans were less able to correct mistakes made by interpretable models, suggesting that we can be lulled into a false sense of comfort when the model appears understandable. 
These findings agree with older human-automation trust studies, where knowing why an automation system erred increased trust in it, even when this trust was unwarranted \cite{DZINDOLET2003697}. 
While these issues do not preclude the use of interpretability et al., they support approaches to trust centered around understandable behavior certificates formed by handling interpretability with care.



\subsection{The utility of interpretability beyond trust}

Lastly, we investigate three areas that are often touted as major motivations for interpretability et al. In each, we find that while interpretability can be helpful, is not always necessary.

\textbf{Legal regulatory compliance and interpretability.}
In some cases, explaining model behavior is plainly required under legal policies. Under the U.S. Equal Credit Opportunity Act, creditors must identify the main factors impacting a credit score when denying credit. 

In other cases, however, legal requirements of explanation may be less clear, or very limited \cite{gdpr}. While the E.U's GDPR's "right to explanation" has often been used as motivation driving research into interpretable and explainable ML, the policy mandates this only for decisions "based solely on automated processing", which corresponds narrowly to the highest level on the 10-point Sheridan-Verplanck automation scale \cite{sheridanverplanck}. The GDPR policy has instead been described as "right to be informed" on the significance and the envisaged consequences of automated decision-making systems and the "right not to be subject to automated decision-making", to demand human involvement in decisions. 
These rights align with our recommendations on building and communicating about trust in AI while not requiring interpretability or explanations. 
In summary, explanations are certainly essential when legally required, but at the present moment, policy demands requiring widespread adoption of interpretability methods and research may be overstated.

\textbf{Model debugging without interpretability.}
Model debugging, and development in general, can be understood primarily in terms of model behavior. Bugs are undesirable model input-output behavior, and can be detected with level 2 model access; interpretability can be of help here, but is not strictly necessary. The act of debugging or fixing a bug regards reasoning about the impact of changing the training dataset, loss function, or architecture on model behavior. 

Practical approaches for finding bugs include sensitivity analysis ("what-if" analysis) including random attacks, finding adversarial examples, OOD evaluation and analyzing and explaining error residuals \cite{hall}. 
Comparing the model's reasoning process to human expectations is a powerful tool for finding bugs, but less broadly applicable than approaches that only require level 2 access. 
Approaches for debugging models include model assertions \cite{kang2020model}, editing models, regularization, data augmentation, data pre-processing, prediction post-processing, and anomaly detection \cite{hall}. Among these methods, the vast majority are compatible with level 2 access. 
While some models can be directly edited, such as decision trees, this approach is largely obviated by the widespread adoption of finetuning models. 
The differentiability of deep models is a key attribute that enables us to "edit" trained models on a higher level of abstraction: instead of editing individual decision branches, we only need to specify loss functions or regularizers and additional data.

Modifying model behavior can also be interpreted as black-box hyperparameter optimization which can be treated with Bayesian optimization techniques and hypernets \cite{pmlr-v108-lorraine20a}. In the appendix, we give perhaps the simplest algorithm (A\ref{alg:debug}) for improving model behavior by random search in model design space, which can be viewed as an evolutionary algorithm executed manually by developers
. Despite its simplicity, this algorithm broadly captures the production of knowledge in ML research and popularization of new model architectures by "graduate student descent"
 \cite{gencoglu2019hark}. 

\textbf{Scientific discovery without interpretability.}
A classic approach to scientific discovery is model selection. Models encode assumptions and hypotheses, and guided by Occam's razor, the evidence for and against these assumptions is weighted by each model's fit to experimental data over time \cite{ding, shenthesis}. Simple models can directly encode hypotheses as quantitative variables, while more other models can encode hypotheses qualitatively while learning some quantitative aspects from data, with fit evaluated on held-out data (appendix; A\ref{alg:science}) \cite{shen, shenthesis}. Such models include universal differential equations \cite{rackauckas} and other approaches for integrating prior domain knowledge into models \cite{willard2021integrating}. Notably, this process only requires level 2 model access.

The time-tested efficacy of model selection contrasts with the hopes of scientific discovery by interpreting deep models, which are challenged by pervasive shortcut learning and the difficulties in weighing evidence of diverse data modalities with proper uncertainties to the degree that human scientists do. Interpreting models that have succumbed to shortcut learning merely yields shortcuts as insights that can fail to generalize to unseen data, limiting their scientific value. To address this, it is common to constrain the model, often by incorporating prior knowledge, which leads us back towards model selection. 


\section{Discussion}

By clarifying processes of human-AI trust, we have questioned the perceived centrality of interpretability to trust in popular discourse, while advocating for approaches such as evaluating models on held-out data and tasks beyond training losses that are representative of real-world applications \cite{bioimage}, robustness testing \cite{djolonga2020robustness, robustness}, adapting software testing principles to ML \cite{softwaretesting}, and reporting model cards \cite{modelcards} and supplier declarations of conformity \cite{varshney}. 
We emphasized the promise of research towards new behavior certificates that enable new model behaviors, and are more correct, relevant to trust contracts, and understandable. 
Looking forward, it is critical to advance trust with the right mindsets and tools, as human trust governs the extent and manner of AI's impact on our future.




\newpage

\begin{acks}
  We thank Anastasiya Belyaeva and Sebastian Gehrmann for insightful discussions.
\end{acks}

\bibliographystyle{ACM-Reference-Format}
\bibliography{trustworthy}

\appendix

\section{Research Methods}

\subsection{Algorithms}
\label{section:a1}

Here, we present two simple algorithms that demonstrate processes for model debugging and scientific discovery without interpretability.

\noindent
\begin{minipage}[t]{0.48\textwidth}
    \vspace{-5pt}
    \small
    \begin{algorithm}[H]                
      \SetCustomAlgoRuledWidth{0.48\textwidth}  
      \setstretch{1}

      \SetKwInOut{Input}{Input}
      \SetKwInOut{Output}{Output}
    
      \Input{
          Initial hyperparameters $\lambda_{t}$, model $f_\lambda(\theta)$ with learnable parameters $\theta$, observed data $X$, data augmentation function $\phi_\lambda$, training procedure $\texttt{train}_\lambda(X, f) \rightarrow \theta^*$, contract evaluation function $\texttt{contract\_eval}$, mutation function $\texttt{mutate}(\lambda)$ that samples around a given $\lambda$; number of mutations $n$
        }
        \Output{$\lambda_{t+1}$, hyperparameters with possibly improved model performance}
    
      \For{$i = 1, ..., n$}
        {
          $\lambda_{i} = \texttt{mutate}(\lambda_t)$ \\
          $\theta_i^* = \texttt{train}_{\lambda_i}(\phi_\lambda(X), f_{\lambda}(\theta_{\texttt{init}}))$ \\
          $s_i = \texttt{contract\_eval}(f_{\lambda_i}(\theta^*)) $ \\
        }
      \textbf{return} $\lambda_{i^*}$ with maximum $s_i$ \\
      \caption{Model development with lv. 2 access via an evolutionary algorithm}
      \label{alg:debug}

    \end{algorithm}
\end{minipage}
\hfill
\begin{minipage}[t]{0.48\textwidth}
  \vspace{-5pt}
  \small
  \begin{algorithm}[H]                
    \SetCustomAlgoRuledWidth{0.48\textwidth}  
    \setstretch{1}

    \SetKwInOut{Input}{Input}
    \SetKwInOut{Output}{Output}

    \Input{
        A set of $n$ models $f_i(\theta)$ representing distinct scientific hypotheses with learnable parameters $\theta$, indexed by $i$; data $X_\texttt{train}$, $X_\texttt{test}$; procedure $\texttt{train}(X, f) \rightarrow \theta^*$; prior belief distribution over hypotheses and models $p(f_i(\theta))$, e.g., Occam's razor
      }
      \Output{$p(f_i|X)$. Updated hypothesis beliefs given data.}

    \For{$i = 1, ..., n$}
      {
        $\theta_i^* = \texttt{train}(X_{\texttt{train}}, f_i(\theta_{\texttt{init}}))$ \\
        $f_i^* = f_i(\theta^*)$ \\
      }
    \textbf{return} $p(f_i|X) = \frac{p(X_{\texttt{test}}|f_i^*) p(f_i^*)}{ \sum_{j} p(X_{\texttt{test}}|f_j^*) p(f_j^*) }$
    \caption{Scientific discovery with lv. 2 access by Bayes rule}
    \label{alg:science}
  \end{algorithm}
\end{minipage}

\end{document}